# Dates Fruit Disease Recognition using Machine Learning

Ghazanfar Latif
*Computer Science Department*
*Prince Mohammad Bin Fahd University*
Khobar, Saudi
Arabiaglatif@pmu.edu.sa

Jaafar Alghazo
*Computer Engineering Department*
*Virginia Military Institute*
Lexington, VA, U.S.A
alghazojm@vmi.edu

Ghassen Ben Brahim
line 2: *Computer Science Department*
line 3: *Prince Mohammad Bin Fahd University*
Khobar, Saudi Arabia
gbrahim@pmu.edu.sa

Khalid Alnujaidi
*Computer Science Department*
*Prince Mohammad Bin Fahd University*
Khobar, Saudi Arabia
202002530@pmu.edu.sa

*Abstract*— Many countries such as Saudi Arabia, Morocco and Tunisia are among the top exporters and consumers of palm date fruits. Date fruit production plays a major role in the economies of the date fruit exporting countries. Date fruits are susceptible to disease just like any fruit and early detection and intervention can end up saving the produce. However, with the vast farming lands, it is nearly impossible for farmers to observe date trees on a frequent basis for early disease detection. In addition, even with human observation the process is prone to human error and increases the date fruit cost. With the recent advances in computer vision, machine learning, drone technology, and other technologies; an integrated solution can be proposed for the automatic detection of date fruit disease. In this paper, a hybrid features based method with the standard classifiers is proposed based on the extraction of L*a*b color features, statistical features, and Discrete Wavelet Transform (DWT) texture features for the early detection and classification of date fruit disease. A dataset was developed for this work consisting of 871 images divided into the following classes; Healthy date, Initial stage of disease, Malnourished date, and Parasite infected. The extracted features were input to common classifiers such as the Random Forest (RF), Multilayer Perceptron (MLP), Naïve Bayes (NB), and Fuzzy Decision Trees (FDT). The highest average accuracy was achieved when combining the L*a*b, Statistical, and DWT Features.

## I. INTRODUCTION

Throughout the last couple of decades, many advancements have been made in computer vision methods. This has led to a drastic increase in the performance of machine learning processes and the creation of better Artificial Intelligence (AI) solutions. Artificial intelligence has become a crucial technology in our societies, it has been directly involved in the increase of global quality of life. We are now seeing the use of AI and image processing techniques increasingly being implemented in industrial processing, medical imaging, agriculture, and so many more places [1].

Saudi Arabia is one of the leading countries in the production and consumption of palm dates. Date fruits play significant roles both economically and culturally within the country. The date palm (Phoenix dactylifera) is a native plant to the Middle East and North African (MENA) parts of the world. It holds a special place for the people of the MENA region since ancient times. This is mostly due to its special properties of being one of the very few plant species thatcan grow in the hot desert climate and provide such a highly nutritious fruit [2]. The Kingdom is the leading country inthe world in regard to average per capita consumption [3]. In the Qassim region of the country, during cultivation season a huge date festival is held, where people from all neighboring regions visit and indulge in the best quality dates. NicknamingQassim as the date city. It is appearing that date fruit holds cultural significance.

The date fruit plays a big role in the economy of the country as well. The kingdom of Saudi Arabia is home to an estimated 30 million date palm trees spread over more than 100 thousand hectares [3]. This is part of a drastic increase in production over the last couple of decades, with an increase ofapproximately 86% in production between 2000-2010. Thoughwith this big cultivation increase, the rate at which the fruit is being processed during their harvesting season is very slow; Aslittle as only 10% of an increase in processing speed has been managed [4]. This is mostly due to the fact that the process is still heavily dependent on human labor. From separating different types to then checking the quality of the harvest, to the manual packaging. Therefore, it is clear that it is crucial to start implementing the use of AI technologies to help boost the performance of processing.

This research will present different approaches that can be helpful in the automation of the classification and detection of date fruits and their diseases using machine learning and AI techniques. It is very clear that the help of technology in the agricultural sector has helped in achieving remarkable growth. However, there are still many challenges that the agricultural domain faces. These challenges are present in both the fields and the factories. Problems such as diseased plants, pest infestations, processing big supplies, big data, and so on, are repetitive and tedious to humans and there is a lot of room for error. Therefore, it should be a priority to start to implement newer and smarter technologies to help prevent crop losses and increase productivity within factories [5].

The importance of the date fruits for country economies such as Saudi Arabia and Tunisia among others has a huge economic impact on these countries. The manual labor of processing these date fruits for exportation is prone to human errors and can make the date fruits more expensive. In addition, early detection of date fruit can definitely assist in early intervention to save produce. With the vast farming lands of date trees, it is next to impossible to have farmers inspect each date tree on a daily basis or even on a weekly basis. Thus, machine learning algorithms fitted on drone technology can be able to assist in this task and be vital in the early detection and early intervention of date tree/fruit disease. Also, Machine learning can be very helpful in processing date fruits once they are picked eliminating human errors and reducing costs.

The aim of this research is to build on the idea of using artificial intelligence in agriculture. Specifically, this research will look to implement artificial intelligence technologies for the purpose of detecting different date fruit diseases in the preharvest stage. The same algorithms can then be modified in future research for the processing of date fruit in the factories prior to market.

The rest of the paper is organized as follows: first, a review of related literature is conducted and presented. This is followed the methodology which includes a comprehensive description of the dataset, feature extraction, and the various classifiers used. The results are presented in a separate section, and finally concluded by discussing the results and possible future development.

## II. RECENT STUDIES

In [6] multiple classification methods were compared: Support Vector Machine (SVM), Neural Network, Decision Tree (DT), and Random Forest (RF) model. A total of four classes composed of the four common date fruit types in Saudi Arabia were used. The dataset was made of 325 images, though the quality of the data was lower due to the fact the images were gathered from different online sources, each of which was captured with differing cameras. The method used for extracting features was a Three Level DWT for texture features and converting the image to an RGB color scheme for color feature extraction. The highest accuracy reported was using the SVM with 73.8% outperforming other classification methods.

With an additional increase in the quality of imagery better performance is expected. In [7], much higher accuracy of 97.26% was achieved using a Neural Network, outperforming the support vector machine, and K-nearest neighbor. This can be attributed to the process of using a fixed quality of images by hardware setup for individually taking the images of the date fruit. As well as growing the dataset for each of the individual classes. The dataset for this research consisted of 600 date fruit samples for a total of six classes each representing one type of the most common date fruit types in Oman, each class contained 100 images. A total of 19 features based on color, shape, size, and texture features were extracted and used for the classification.

In [8], the approach of using Deep Learning was used in classifying the date fruit. The Deep Learning technique uses several hidden layers and a total of 100 epochs to create a framework that autonomously detects the different types of date fruit. Three types of date fruit were used for this research: Aseel, Karbalain, and Kupro. The dataset used to run the experiment consisted of 500 images separated among the three classes. Thresholding of the date image was applied to then extract the histogram values, color features, size features, and shape features. The created framework using Deep Learning was able to give an accuracy of 89.2%.

In [9], the research aimed in using three different machine learning methods in the detection of the date fruit. The experiments were executed on seven different varieties of dates. The dataset was composed of 898 total images, the acquisition of the data was through a custom system of hardware setup with the purpose of taking individual images of the samples. Image thresholding was done through the Otsu method. A total of 34 morphological features were extracted. For classification, Logistic Regression (LR) Analysis, Artificial Neural Network (ANN), and Stacking of each module were used. Both classification methods achieved over 90% accuracy while stacking the two modules produce a 92.8% accuracy.

In [10], the study included a total of 11 types of date fruit types with a total of 660 images divided across three classification methods: SVM, ANN, and Probabilistic Neural Network (PNN). The use of Pruning Local Distance-based Outlier Factor method was used to prune out outlier samples. The feature extracted were the color, texture, and size. For the color features, the color channels from the RGB space were quantized into 12 bins. For texture extraction, they used Gray-Level Co-occurrence Matrix. As for size features, the principal axes method was used. With the use of a 3-fold cross-validation technique, an accuracy of 98.65% was achieved.

The focus of the work in [11] was to automate the inspection of the external quality of Khalas date. 566 images of two classes were used, good quality, and sugaring dates. For the classification, the bag of feature (BOG) method was used. The training stage consisted of Key point detection using the grid method, Feature extraction using the SURF detector method, Creation of a directory using K-Means clustering, and Vector Quantization for each of the classes. Results showed that with the use of the SVM method, an average accuracy of 99% classification with 133 features and 50 clusters was achieved.

In [12], the focus was on evaluating Convolutional Neural Networks in determining the ripeness of Medjool Dates. Eight different convolutional neural networks (CNN) architectures: VGG-16, VGG-19, ResNet-50, ResNet-101, ResNet-152, AlexNet, Inception V3, and CNN were evaluated. The comparison was based on the accuracy of detecting the maturity rate of Medjool Dates. A total of 1002 images were used as different ripeness levels. The results here showed that the VGG-19 architecture performed the best with a 99.32% accuracy rate.

In [13] researchers worked on using Deep Learning for the classification of different apple fruit diseases. The classification was based on four classes, Healthy, Blotch, Rot, and Scab. Experiments were done with different Deep Convolutional Neural Networks (DCNN). The original data contained 319 images in total for all of the classes. Since deep learning requires large data samples, the researchers have created an additional 4,000 synthetic images using the Deep Convolutional Generative Adversarial Network architect. Results showed that the use of a DCNN based on Adam optimizer achieved an average accuracy of 99.99% detection rate.

In [14], the authors looked at the classification of apple fruits as well. The research worked with four classes, Rot, Scab, Blotch, and Normal apples. The methods used in these experiments were image segmentation, and extraction of color, texture, and shape features. With the combination of features, a Multi-class Support Vector Machine (MSVM) was used for the training and classification. The dataset consisted of 280 total images equally divided between the four classes. The results showed with this approach they were able to achieve an accuracy of 96%.

In [15], the research focused on the automatic detection of citrus fruit disease using machine learning. The methods used for the experiments were K-means clustering, ANN, and SVM. Four features are extracted from the images. Color features using RGB values. Texture features morphology, and structure of holes using the 2-D deviation of a grey level. The dataset consisted of five classes each of a different common citrus disease. The result of the experiments showed that the SVM achieved an accuracy of 93.12%. The research in [16] was based on the grading of mango fruit based on the CIELAB color model. The experiments were conducted with 100 mango samples, 60 of which were healthy and 40 diseased. Feature extraction consisted of color based on the CIELAB color channel, and size feature using a diameter equation based on the binary image of mango. With the Dominant Density Range method, the purposed methods achieved an average accuracy of 92.37%. In [17], the authors propose the use of a Deep Convolutional Neural network (DCNN) with a transfer learning model to accurately detect and classify rice leaf disease. For multiclass classification of 6 different classes, they reported the highest accuracy of 96.08%.

## III. METHODOLOGY

The approach that will be taken in order to do the experiments will consist of several components. There is currently no sufficient existing dataset that is made up of different disease and malformations that are preset in the preharvest stage of the date fruit. The first task was to build the dataset for the experiments, through collection, preprocessing, and labelling. Having clean labeled data, it is possible to run quality experiment. First the features of the images were extracted through several methods covered in the feature extraction section. Lastly, an overview of the classification methods that are going to be used.

### A. Experimental Dataset

The dataset used for this research consists of a total of 871 images which is summarized in Table 1. The dataset makes up four different classes: Healthy date, Initial stage of disease, Malnourished date, and Parasite infected.

TABLE 1. EXPERIMENTAL DATASET SUMMARY WITH SAMPLE IMAGES

| Class | Images | Sample 1 | Sample 2 |
|---|---|---|---|
| *Healthy* | 227 | 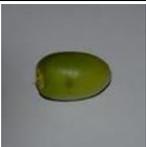 | 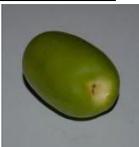 |
| *Initial Stage of Disease* | 227 | 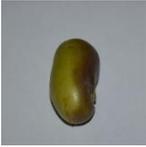 | 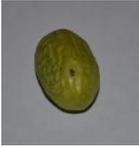 |
| *Malnourished* | 229 | 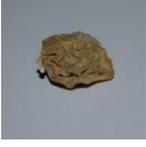 | 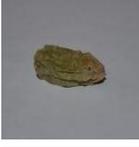 |
| *Parasite Infected* | 254 | 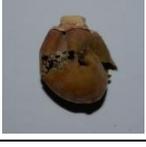 | 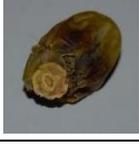 |

### B. Preprocessing of the Images

The original image's dimension was 3072 x 4608. This is very large and bothersome and would make it difficult for a computer to process. So, the first thing done to the images was to decrease their size by 90%, resulting in dimensions 307 x 460. The main focus here was to find the edges of the region of interest (ROI) i.e., the dates. First, the images were turned into grayscale, this is because the focus here is to identify the descriptors of the images as in the edges. Having the images in grayscale help simplifies the algorithm used and some algorithms require the image to be in grayscale.

Basic morphological operations were then run on the gray images. Morphological Operations are a set of operations that process images in regard to their shape. The methods applied to the images were the Erosion method and the Dilation method. These two methods usually are used together. The Erosion of the image consists of eroding the features and expanding the pixels in a way to present disconnection between two objects. Followed by Dilation which removes noise by shrinking the object to have it resemble a proper area. Lastly, Gaussian Blur is used to further reduce any noise in the image.

After the shape of the object within the image is very clear in grayscale. A Canny function is applied to the image to distinguish the edges. With the image being in binary, and the canny applied properly and giving accurate edges, the contours of the image are then found. At this stage, the images have very few contours. Contours are outlines that bound objects that are present in the image. With the help of a functionthat finds the contour with the largest area, the desired ROI coordinates are found within the image. Now the original images can be cropped to the relevant coordinates.

The date fruit samples were gathered from a local farm located in Safwa city in the Eastern Province. The samples were handpicked and then taken back to another location to be photographed. Approximately 40 dates were chosen for each of the four classes. Each sample was photographed individually. The setup consisted of placing each date on a white A4 piece of paper as a background. After placing each of the date samples on the platform they were captured from approximately 5 different angles. The tools used for the proses were Nikon D3100 14.2-megapixel digital camera.

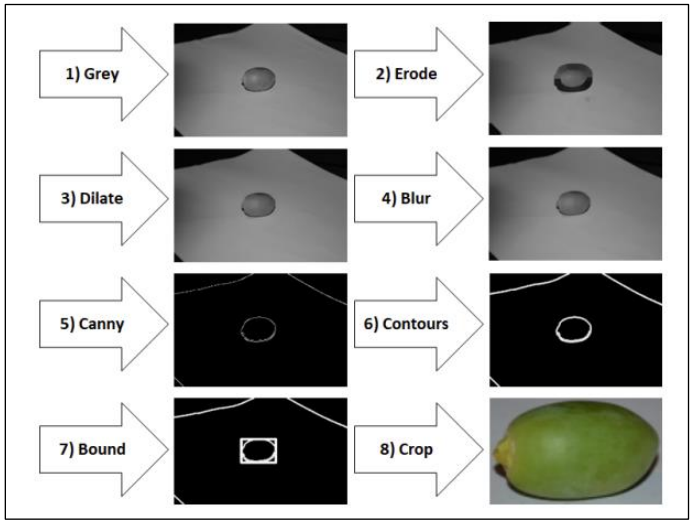

Figure 1. Preprocessing steps for the segmentation of Date Fruit from Images

### C. Feature Extraction

Feature extraction refers to the process of extraction of key feature (or in layman's terms, details) that are present in images. For humans it is simple for us to tell apart two different objects, however, computers only understand 0's and 1's. The extraction of features creates key tables of numerical values that are compressed in a way that assists in the computations used elsewhere. These key features of the different classes are then fed into a classification model. The features that are useful to extract out of the image are: the color of the object, the shape of the object, and the colors of the object. There are many methods used for feature extraction, in this paper 3 will be used: L*a*b color features, Discrete Wavelet Features, and Statistical features.

*1) L*a*b Color Features:* L*a*b or commonly named CIELAB is a color space similar to red, green, and blue (RGB). The L*a*b* color space however consists of three different layers; L* luminosity layer resampling where the color falls between white-black axis; layer b* resembling chromaticity of the color between red-green axis, and layer a* resembling chromaticity of the color between blue-yellow axis. It is known for its very close resemblance to the human perception of colors [18].

Images are generally represented RGB values, in order to be able to extract features using L*a*b* the image's colors must be converted respectively. The is no direct formula to convert from RGB to L*a*b*. So, the process consists of two steps. First, the color is converted to an intermediate color space, this being either sRGB or CIEXYZ. The transformation of the colors from RGB to CIEXYZ can be done through equations (1) and (2) [19]:

$$\begin{bmatrix} X \\ Y \\ Z \end{bmatrix} = \begin{bmatrix} 0.607 & 0.174 & 0.200 \\ 0.299 & 0.587 & 0.114 \\ 0.000 & 0.066 & 1.116 \end{bmatrix} \begin{bmatrix} R \\ G \\ B \end{bmatrix} \quad (1)$$

$$L* = 116\left(\sqrt[3]{\frac{y}{y_0}}\right) - 16$$

$$a* = 500\left(\sqrt[3]{\frac{x}{x_0}} - \sqrt[3]{\frac{y}{y_0}}\right) \quad (2)$$

$$a* = 200\left(\sqrt[3]{\frac{y}{y_0}} - \sqrt[3]{\frac{z}{z_0}}\right)$$

Where y y0 >0.01, x x0 >0.01, and z z0 >0.01. (X0, Y0, Z0) are the X, Y, and Z values for standard white color. To further get the values in L*a*b* scale, the following Equation 3 is applied

$$\Delta E_{ab} = \sqrt{(\Delta L*)^2 + (\Delta a*)^2 + (\Delta b*)^2} \quad (1)$$

*2) Discrete Wavelet Transform (DWT):* is one of the most relevant methods used in computer vision and machine learning for the purpose of object detection and classification. It is used in many applications in multiple fields and industries. DWT works by transferring signals into basic functions known as wavelets [20], where each wavelet holds unique values known as wavelet coefficients. These wavelet coefficients are then a representation of feature vectors used to identify the object. The DWT decomposes an image into four sub-images or four subsets: LL, LH, HL, and HH, each holding a special property of low and high frequencies. Detailed coefficients are represented in HH, LH, and HL, and approximated coefficients are presented in LL. These distinguished coefficients are then fed to a classification module to be used for the analysis in order to detect and classify the object in the image.

*3) Statistical Features:* can be seen as a way to group the recurrent patterns that are represented in features that can be compacted into specific representations. This operation is approached by using a special equation in the field of statistics. There are two levels of extracting features using the statis- tical method. First-order statistics include Mean, Variance, Skewness, kurtosis, energy, and entropy [21]. The second one includes second-moment energy, correlation, inertia, absolute value, inverse difference, entropy, and maximum probability. In this article total of 12 statistical features are used along with L*a*b and DWT Features.

*D. Classification*

Classification is the process in which an AI model makes a prediction. The classification process takes in various inputs and proceeds to run it through algorithms to do a set of calculation in order to make a prediction. This is done through having previously trained the same algorithms and having clear indication for the outputs to represent certain classes. The measure of performance in which a model can classify depends on accuracy of the model, the complexity of the problem, and quality of the data. In this paper several classification methods will be used: Random Forest, Fuzzy Decision Tree, Multilayer Perceptron, and Bayesian Network.

*1) Random Forest (RF):* is an ensemble method of classification. The RF algorithm works through the creation of a collection of multiple variations of Decision Trees (DT). Where each DT is trained on a bootstrap set of data, and the classification of the RF depends on the majority or average of the output of the many DTs. The RF is empowered through the use of DT, by reducing the high variance it has [22]. The RF is made up of n number of individual DT with user-defined parameter, where the accuracy of the overall RF is significantly improved as the number of trees increases. Each of the DT modules is given a set of data "with replacement" as no two trees in sequence will be fed the same sample of data, this is to decrease the lower variance of the overall RF. In the training of the RF, the algorithm determines the node to split. This is typically the square root of the number of features in the dataset.

*2) Fuzzy Decision Trees:* Fuzzy logic is when a set of values or outcomes can be reprehensive of a given value. Similar to how humans are able extract context and classify objects in vague or ambiguous matter, Fuzzy Decision Trees (FDT) to try mimic this. Decision Trees (DT) are a very popular learning and classification method. FDT build up on the thorough DT structure by adding fuzzy representation to assist to furthers the accuracy. The FDT can better process noisy symbolic, and numerical data. This is through fuzzy restriction that are applied to be able to deal with continuous domains of values by using fuzzy membership functions [23].

*3) Multilayer Perceptron:* The Multilayer Perceptron (MLP) is one of the basic and most popular Artificial Neural Networks (ANN). It is considered to be a forward ANN, as the signals of the network travel through the network in one direction. The training of the MLP can be supervised with the machine learning technique of backpropagation. MLP structure consists mainly of at least three different layers: Input, hidden layer, and output layer. The number of input layers being user defined, taking as many arguments of the objects or classes to train the network. Second comes the hidden layer, so-called because it does not have direct contact with either input or output of the environment [24]. The hidden layers are where the training of the module take place. It does this through a backpropagation algorithm. The main idea of this process is to minimize the error function. First, all the network connections are initialized with a small range of random values, to begin with to be used as the weight between the neurons or nodes, which later keep getting adjusted using activation functions until the desired result is achieved. Multiple activation functions can be used within MLP, the most common one being the Sigmoid activation function.

*4) Bayesian Network:* A Bayesian Network (BN) is a probabilistic graphical module that is used for classification and regression. It is more specifically a type of graphical model known as a directed acyclic graph (DAG). This type of graph shows the relationship between nodes and their parents, as a one-way relationship where they point in a particular direction. These relationships are to be presentative of the distribution of conditional probability. The Bayesian network can be of use in the inferencing and learning of data. Two main components of learning in BN are structure learning, and parameter learning. Structure learning works on discovering the best relationship representation of the data within the DAG. The use of two popular methods is implemented to determine the structure of the DAG; the DAG search algorithm, and the K2 algorithm. These two algorithms work on giving the Bayesian score. Parameter learning focuses on identifying conditional probability distributions. There are multiple BN classifiers, however using a supervised learning module the focus here will be on the naive Bayes classifier, as it has been very successful in a broad range of fields [25].

## IV. RESULTS

After the text edit has been completed, the paper is ready for the template. Duplicate the template file by using the Save As command, and use the naming convention prescribed by your conference for the name of your paper. In this newly created file, highlight all of the contents and import your prepared text file. You are now ready to style your paper; use the scroll down window on the left of the MS Word Formatting toolbar.

The same experiment was repeated using the statistical features only. The results are shown in Table 3. The highest average accuracy in this case was achieved using the MLP classifier. The highest average accuracy achieved was 79.31% with a Precision, Recall, and F1-measure of 79.3%, 79.3%, and 79.10%, respectively. It is observed that the lowest accuracy is reported when using the BN classifier. The lowest average accuracy is 79.31% using the statistical features only with the BN classifier.

TABLE 2. EXPERIMENTAL RESULTS FOR DIFFERENT CLASSIFIERS USING L*A*B FEATURES

| Classifier | Accuracy | Precision | Recall | F1-Measure |
|---|---|---|---|---|
| RF | **81.61%** | **81.30%** | **81.36%** | **81.30%** |
| BN | 77.01% | 76.20% | 77.00% | 76.60% |
| MLP | 81.03% | 81.20% | 81.00% | 81.10% |
| FDT | 70.69% | 71.10% | 70.70% | 70.30% |

Table 3. EXPERIMENTAL RESULTS FOR DIFFERENT CLASSIFIERS USING STATISTICAL FEATURES

| Classifier | Accuracy | Precision | Recall | F1-Measure |
|------------|----------|-----------|--------|------------|
| RF | 72.41% | 72.10% | 72.40% | 72.00% |
| BN | 64.37% | 65.50% | 64.40% | 64.10% |
| MLP | **79.31%** | **79.30%** | **79.30%** | **79.10%** |
| FDT | 64.94% | 65.30% | 64.90% | 65.10% |

Results of combining both the L*a*b color features and the statistical feature and inputting them the common classifier are shown in Table 4. In this case, the highest average accuracy was achieved using the MLP classifier. The highest average accuracy in this case was 88.51% with a Precision, Recall,and F1-measure of 89.2%, 88.5%, and 88.6%, respectively. FDT again is the low performer in this case and the lowest average accuracy is 74.71% using the combined L*a*b color features and the statistical features input to the FDT classifier.

Table 4. EXPERIMENTAL RESULTS FOR DIFFERENT CLASSIFIERS USING COMBINED L*A*B AND STATISTICAL FEATURES

| Classifier | Accuracy | Precision | Recall | F1-Measure |
|------------|----------|-----------|--------|------------|
| RF | 83.33% | 83.40% | 83.30% | 82.80% |
| BN | 79.89% | 79.40% | 79.90% | 79.60% |
| MLP | **88.51%** | **89.20%** | **88.50%** | **88.60%** |
| FDT | 74.71% | 74.90% | 74.70% | 74.60% |

Through careful observation of the Tables 1-4, we notice that combining the L*a*b, statistical, and DWT features has shown signification improvement in the classification com- pared to the use of L*a*b color features alone with an increase of 8.05%, the use of statistical features alone with an increase of 10.35%, and even the combined L*a*b and statistical features with an increase of 1.15%.

Figure 2 shows the comparison of classification using the combined L*a*b, statistical, and DWT features. It can be seen that the best metric results are achieved using the MLP classifier. Even though healthy dates are classified correctly in other classifiers, however, when observing the results for the initial stage of disease (orange bar), malnourished (gray bar), and parasite infection (yellow bar), it is immediately noticed that the best results are achieved using the MLP classifier.

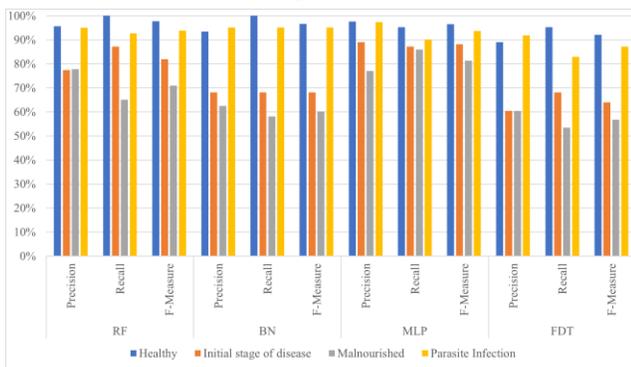

FIGURE 2. COMPARISON OF DIFFERENT CLASS RESULTS USING THE COMBINED L*A*B, STATISTICAL, AND DWT FEATURES

## V. CONCLUTION

Date fruit is important in many parts of the world and is part of their culture and diet. Among the top exporters of the fruit date are Saudi Arabia and Tunisia. This means that it is part of the country's economic well-being. Like other agricultural products, date palm trees are susceptible to some diseases. The early detection and intervention of palm date fruit disease are vital in saving the produce and stopping its spread among the surrounding palm trees. This requires that frequent checks of the trees and date fruits be done. However, checking vast farmlands is almost next to impossible and would probably require farmers to randomly check on portions of the trees which means that they might miss the early detection of the disease. Checking all the trees and all the farmland by farmers on a frequent basis is next to impossible. Even if it were possible, this would increase labor costs and end up increasing the price of date fruits for the end consumer. With the new developments in machine learning, the Internet of Things (IoT), cloud computing, drone technology, and others, a device that integrates these technologies can be developed to capture images, process them and detect whether they are healthy or not. The main part of such a system would really depend on developing a machine learning algorithm that can accurately detect and classify date fruit disease. It should be noted that machine learning is widely used and researched nowadays not only in the agricultural fields but also in many other fields. For example, in Medicine research is developing machine learning algorithms for diagnosis [32], and in Smart Cities, machine learning algorithms are used Smart Applications [33]. Machine learning is used for many applications in all walks of life.

In this paper, the aim is to develop a machine learning algorithm that is able to accurately detect and classify four classes: Health, Initial stage of disease, Malnourished date, and Parasite infected. Since our concentration was on date fruit diseases in Saudi Arabia, we developed a dataset from images of date fruits taken from Saudi Arabian Farms. The dataset consists of 871 images. We also propose extracting the L*a*b color features, statistical features, and DWT features. Experiments were performed on the combination of features input to common classifiers namely RF, BN, MLP, and FDT. It was found that the highest accuracy is achieved when combining all three feature categories and when using theMLP classifier. The highest average accuracy in this was 89.66% with Precision, Recall, and F1-Score of 90.2%, 89.7%, and 89.8%, respectively. Compared with similar methods fromthe extant literature, the proposed method showed superior performance. It should be noted however that the dataset was developed for this paper and thus direct comparison is not possible.

In the future, we plan to continue the research on developing a larger date fruit dataset and also improve on the machine learning algorithms so as to reach an optimal solution for date fruit disease detection. We also plan to utilize the optimal solution and integrate it with IoT and Drone Technology to test it in the farmlands.